\def\year{2016}\relax
\documentclass[letterpaper]{article}
\usepackage{aaai16}
\usepackage{mathptmx}
\usepackage{times}
\usepackage{helvet}
\usepackage{courier}
\usepackage{algorithm}
\usepackage[noend]{algorithmic}
\usepackage{graphicx}
\usepackage{xcolor}
\usepackage{float}
\usepackage{amsmath}
\usepackage{amssymb}
\usepackage{amsthm}
\usepackage{subfig}
\usepackage{booktabs}
\usepackage{natbib}

\newtheorem{exampl}{Example}
\newenvironment{example}{\begin{exampl}\em}{\end{exampl}}
\newtheorem{defn}{Definition}

\begin{document}

\title{Why is Compiling Lifted Inference into a Low-Level Language so Effective?\thanks{In IJCAI-16 Statistical Relational AI Workshop.}}
\author{
Seyed Mehran Kazemi \and David Poole\\
The University of British Columbia\\
Vancouver, BC, V6T 1Z4\\
\{smkazemi, poole\}@cs.ubc.ca
}
\maketitle
\begin{abstract}
First-order knowledge compilation techniques have proven efficient for lifted inference. They compile a relational probability model into a target circuit on which many inference queries can be answered efficiently. Early methods used data structures as their target circuit. In our KR-2016 paper, we showed that compiling to a low-level program instead of a data structure offers orders of magnitude speedup, resulting in the state-of-the-art lifted inference technique. In this paper, we conduct experiments to address two questions regarding our KR-2016 results: 1- does the speedup come from more efficient compilation or more efficient reasoning with the target circuit?, and 2- why are low-level programs more efficient target circuits than data structures? 
\end{abstract}

Probabilistic relational models \citep{getoor2007introduction,StarAI-Book} (PRMs), or template-based models \citep{Koller:2009}, are extensions of Markov and belief networks \citep{Pearl:1988} that allow modelling of the dependencies among relations of individuals, and use a form of exchangeability: individuals about which there exists the same information are treated identically. The promise of lifted probabilistic inference \citep{Poole:2003,kersting2012lifted} is to carry out probabilistic inference for a PRM without needing to reason about each individual separately (grounding out the representation) by instead exploiting exchangeability to count undistinguished individuals.

The problem of lifted probabilistic inference was first explicitly proposed by \citet{Poole:2003}, who formulated the problem in terms of parametrized random variables, introduced the use of splitting to complement unification, the parametric factor (parfactor) representation of intermediate results, and an algorithm for summing out parametrized random variables and multiplying parfactors in a lifted manner. This work was advanced by the introduction of counting formulae, the development of counting elimination algorithms, and lifting the aggregator functions \citep{De:2005,Milch:2008,Kisynski:2009a,Choi:2011,Taghipour:2014}. The main problem with these proposals is that they are based on variable elimination. Variable elimination \citep{ZhangPoole:1994} (VE) is a dynamic programming approach which requires a representation of the intermediate results, and the current representations for such results are not closed under all operations used for inference.

An alternative to VE is to use search-based methods based on conditioning such as recursive conditioning \citep{Darwiche:2001}, AND-OR search \citep{Dechter:2007} and other related works (e.g., \citet{Bacchus:2009}). While for lifted inference these methods require propositionalization in the same cases VE does, the advantage of these methods is that conditioning simplifies the representations rather than complicating them, and these methods exploit context specific independence \citep{Boutilier:1996} and determinism. The use of lifted search-based inference was proposed by \citet{Jha:2010}, \citet{PTP} and \citet{Poole:2011}. These methods take as input a probabilistic relational model, a query, and some observations, and output the probability of the query given the observations. 

\citet{Van:2011} and \citet{LRC2CPP} follow a knowledge compilation approach to lifted inference by evaluating a search-based lifted inference algorithm symbolically (instead of numerically) and extracting a target circuit on which many inference queries can be efficiently answered. While the target circuit used by \citet{Van:2011} is a data structure, \citet{LRC2CPP} showed that using a low-level program as a target circuit is more efficient and results in orders of magnitude speedup. In a simultaneous work, \citet{SWIFT} showed that compiling to a low-level program is effective for BLOG \citep{BLOG} and offers substantial speedup.

Two issues remained unanswered in \citet{LRC2CPP}'s results: 1- they compared end-to-end (compiling to a target circuit and reasoning with the circuit) run-times of their work with \citet{Van:2011}'s weighted first-order model counting, leaving the question of where exactly the speedup comes from, and 2- the actual reason behind the speedup gained by compiling to a program instead of a data structure remained untested. In this paper, we answer to these two issues. We conduct our experiments on Markov logic networks (MLNs) \citep{Richardson:2006aa} and argue that our results also hold for other representations (e.g., \citep{Jaeger:1997,DeRaedt:2007,suciu2011probabilistic,Kazemi:2014}).

\section{Background and Notations}
A \textbf{population} is a set of \textbf{individuals} (a.k.a. things, entities or objects). The \textbf{population size} is a nonnegative integer indicating the cardinality of the population. A \textbf{logical variable} is written in lower case and is typed with a population. For a logical variable $x$, we let $\Delta_x$ and $|\Delta_x|$ represent the population associated with $x$ and its cardinality respectively. A lower case letter in bold represents a tuple of logical variables.
Constants, denoting individuals, are written starting with an upper-case letter. 
A \textbf{term} is a logical variable or a constant.

A \textbf{parametrized random variable (PRV)} consists of a k-ary predicate $R$ and $k$ terms $t_i$ and is represented as $R(t_1,\dots, t_k)$. 
If every $t_i$ is a constant, the PRV corresponds to a random variable. When k = 0, we omit the parentheses. A \textbf{grounding} of a PRV can be achieved by replacing each of the logical variables with one of the individuals in their domains. 
A \textbf{literal} is an assignment of a value to a PRV. We represent $R(\dots)=True$ and $R(\dots)=False$ by $r(\dots)$ and $\neg r(\dots)$ respectively.
A \textbf{world} is a truth assignment to all groundings of all PRVs. A \textbf{formula} is made up of literals connected with logical connectives (conjunctions and disjunctions).

Following \citet{LRC2CPP}, we represent a \textbf{weighted formula (WF)} as a triple $\left\langle L,F,w \right\rangle$, where $L$ is a set of logical variables, $F$ is a formula whose logical variables are a subset of $L$, and $w$ is a real-valued weight.
For a given WF $\left\langle L,F,w \right\rangle$ and a world $\omega$, we let $\eta(L, F, \omega)$ represent the number of assignments of individuals to the logical variables in $L$ for which $F$ holds in $\omega$. 

\subsection{Markov Logic Networks}
A \textbf{Markov Logic Network (MLN)} consists of a set $\psi$ of WFs. It induces the following probability distribution:
\begin{equation} \label{mln-eq}
Prob(\omega)=\frac{1}{Z}\prod_{\left\langle L,F,w \right\rangle \in \psi} \exp({\eta(L,F,\omega)} * w)
\end{equation}
where $\omega$ is a world and 
\begin{equation}
Z= \sum_{\omega'} (\prod_{\left\langle L,F,w \right\rangle \in \psi} (\exp(\eta(L,F,\omega') * w))
\end{equation}
is the partition (normalization) function. In this paper, we focus on calculating the partition function as many inference queries on MLNs reduce to calculating the partition function. Following  \citet{LRC2CPP}, we assume the formulae in WFs of MLNs are in conjunctive form.

\begin{example} \label{mln-example}
Consider an MLN over three PRVs $R(x, m)$, $S(x, m)$ and $T(x)$, where $\Delta_x=\{X_1,X_2,X_3,X_4,X_5\}$ and $\Delta_m=\{M_1,M_2\}$, with the following WFs:
\begin{center}
$\{ \left\langle \{x,m\}, r(x,m)\wedge s(x,m),1.2 \right\rangle$, $\left\langle \{x,m\}, s(x,m) \wedge t(x), 0.2 \right\rangle \}$
\end{center}
and a world $\omega$ in which $R(X_1,M_1)$, $S(X_1, M_1)$, $S(X_1, M_2)$ and $T(X_1)$ are $True$ and the other ground PRVs are $False$. Then $\eta(\{x,m\}, r(x,m)\wedge s(x,m), \omega)=1$ as there is only one assignment of individuals to $x$ and $m$ ($x=X_1$ and $m=M_1$) for which $r(x,m)\wedge s(x,m)$ holds in $\omega$ and $\eta(\{x,m\}, s(x,m)\wedge t(x), \omega)=2$. Therefore:
\begin{center}
$Pr(\omega) = \frac{1}{Z} (\exp(1 * 1.2) * \exp(2 * 0.2))$
\end{center}
\end{example}

An MLN can be conditioned on a PRV having no logical variables by replacing the PRV in all formulae of WFs with its observed value. Observations on individuals can be handled by a process called shattering. 
\begin{example} \label{mln-obs}
Suppose for the MLN in Example~\ref{mln-example} we observe that $T(X_1)$ and $T(X_2)$ are $True$. Since we have more information about $X_1$ and $X_2$ compared to other individuals in $\Delta_x$, the individuals in $\Delta_x$ are no longer exchangeable. In order to handle this, we create two new logical variables $x_1$ and $x_2$ with $\Delta_{x_1}=\{X_1,X_2\}$ representing the individuals for which we have observed $T$ is $True$, and $\Delta_{x_2}=\{X_3, X_4, X_5\}$ representing the individuals for which we have not observed $T$. Then we create new WFs with our new logical variables as follows:
\begin{center}
$\{ \left\langle \{x_1,m\}, r(x_1,m)\wedge s(x_1,m),1.2 \right\rangle$, $\left\langle \{x_2,m\}, r(x_2,m)\wedge s(x_2,m),1.2 \right\rangle$, $\left\langle \{x_1,m\}, s(x_1,m) \wedge t(x_1), 0.2 \right\rangle$, $\left\langle \{x_2,m\}, s(x_2,m) \wedge t(x_2), 0.2 \right\rangle \}$
\end{center}
Then we replace $t(x_1)$ with $True$. For every logical variable $x$ in the above MLN, the individuals in $\Delta_x$ are now exchangeable. This process is called shattering. We assume our input MLNs have been shattered based on the observations and refer interested readers to \citep{De:2005} for the details.  
\end{example}
An MLN can be also conditioned on some counts: the number of times a PRV with one logical variable is $True$ or $False$. For a PRV $T(x)$, we let $Obs(T(x), i)$ represent a count observation on $T(x)$ indicating $T$ is $True$ for exactly $i$ out of $|\Delta_x|$ individuals.
\begin{example}
Suppose for the MLN in Example~\ref{mln-example} we observe $Obs(T(x), 2)$. We create two new logical variables $x_1$ and $x_2$ representing the subset of $x$ having $T$ $True$ and $False$ respectively, with $|\Delta_{x_1}|=2$ and $|\Delta_{x_2}|=3$.\footnote{The domains can be assigned randomly due to the exchangeability of the individuals.} Then we create new WFs as in Example~\ref{mln-obs}, and replace $t(x_1)$ with $True$ and $t(x_2)$ with $False$.
\end{example}

\section{Search-based Lifted Inference Rules}
There are several rules that are used in search-based lifted inference algorithms. In this section, we describe some of these rules using examples.

\subsection{Lifted Decomposition}
\begin{example} \label{lifted-decomposition}
Consider the MLN in Example~\ref{mln-example}. On the relational level, all PRVs are connected to each other and we only have one connected component. On the grounding, however, for every individual $X_i \in \Delta_x$, we have the following WFs mentioning $X_i$:
\begin{center}
$\{ \left\langle \{\}, r(X_i,M_1)\wedge s(X_i,M_1),1.2 \right\rangle$, \\
$\left\langle \{\}, s(X_i,M_1) \wedge t(X_i), 0.2 \right\rangle$, \\
$\left\langle \{\} , r(X_i,M_2)\wedge s(X_i,M_2),1.2 \right\rangle$,\\
$\left\langle \{\}, s(X_i,M_2) \wedge t(X_i), 0.2 \right\rangle \}$
\end{center}
Notice that the WFs mentioning $X_i$ in the grounding are totally disconnected from the other WFs. Therefore, we have $|\Delta_x|$ connected components that are equivalent up to renaming of the $X_i$ individuals. In this case, $x$ is called a \emph{decomposer} of the network. Given the exchangeability of the individuals, the $Z$ of all these connected components are the same. Therefore, we compute the $Z$ for only one of these connected components, e.g., for an MLN with the following WFs, and raise it to the power of $|\Delta_x|$.
\begin{center}
$\{ \left\langle \{m\}, r(X_1,m)\wedge s(X_1,m),1.2 \right\rangle$, $\left\langle \{m\}, s(X_1,m) \wedge t(X_1), 0.2 \right\rangle \}$
\end{center}
In the above MLN, $x$ has been replaced by one of its individuals. We refer to this as decomposing the MLN on logical variable $x$. While in this example only one logical variable is the decomposer, note that in general a set of logical variables can be the decomposer of an MLN. We point interested readers to \citep{Poole:2011} for a detailed analysis of when a set of logical variables $\mathbf{x}$ is a decomposer of a network.
\end{example}

\subsection{Lifted Case Analysis}
\begin{example} \label{lifted-case-analysis}
Consider the resulting MLN in Example~\ref{lifted-decomposition} after being decomposed on $x$. We can find the partition function for this MLN by a case analysis on the values of a PRV. Suppose we do a case analysis on $S(X_1,m)$. Given that $S(X_1,m)$ represents $|\Delta_m|$ random variables in the grounding, one may think $2^{|\Delta_m|}$ cases must be considered: one for each assignment of values to the random variables. However, the individuals are exchangeable, i.e. we only care about the number of times $S(X_1,m)$ is $True$, not about the individuals that make it $True$. Thus, we only consider $|\Delta_m|+1$ cases with the $i$th case being the case where for $i$ out of $|\Delta_m|$ individuals $S(X_1,m)$ is $True$. We also multiply the $i$th case to $\binom{|\Delta_m|}{i}$ to take into account the number of different assignments to the individuals in $\Delta_m$ for which $S(X_1,m)$ is exactly $i$ times $True$. The case analysis for this PRV will then be: \\
$Z(M)=\sum_{i=0}^{|\Delta_m|} \binom{|\Delta_m|}{i} Z(M | Obs(S(X_1,m), i))$ \\
where $M | Obs(S(X_1, m), i)$ has the following WFs:
\begin{center}
$\{ \left\langle \{m_1\}, True \wedge r(X_1,m_1),1.2 \right\rangle$, \\
$\left\langle \{m_2\}, False \wedge r(X_1,m_2),1.2 \right\rangle$, \\
$\left\langle \{m_1\}, True \wedge t(X_1), 0.2 \right\rangle$, \\
$\left\langle \{m_2\}, False \wedge t(X_1),0.2 \right\rangle \}$
\end{center}
\end{example}

\subsection{Removing False Formulae}
\begin{example} \label{remove-false-formulae}
Consider the resulting MLN in Example~\ref{lifted-case-analysis} after the case analysis on $S(X_1, m)$. The formulae of the second and the fourth WFs are equivalent to $False$ and can be removed from the MLN. However, removing these WFs causes the random variables in $R(X_1, m_2)$ to be totally eliminated from the MLN. To address the effect of these variables, we calculate the $Z$ of the MLN having only the first and third WFs and multiply it by $2^{|\Delta_{m_2}|}$, i.e. the number of possible assignments to the $|\Delta_{m_2}|$ random variables in $R(X_1, m_2)$.
\end{example}

\subsection{Decomposition}
\begin{example} \label{disconnected-components}
Consider the resulting MLN in Example~\ref{remove-false-formulae} after removing the WFs whose formulae are equivalent to $False$. The resulting MLN has two WFs each mentioning different PRVs, i.e. the two WFs are disconnected. In this case, we can find the $Z$ of the first and second formulae (more generally: first and second connected components) separately and return the product.
\end{example}

\subsection{Case Analysis}
\begin{example} \label{case-analysis}
Consider the second connected component of the MLN in Example~\ref{disconnected-components}. The partition function of this MLN can be found by a case analysis on $T(X_1)$ as: $Z(M)=Z(M\mid T(X_1)=True)+Z(M\mid T(X_1)=False)$.
\end{example}

\subsection{Evaluating True Formulae}
\begin{example} \label{evaluating-true-formulae}
Consider the MLN in Example~\ref{case-analysis} conditioned on $T(X_1)=True$. This MLN has one WF:
\begin{center}
$\{ \left\langle \{m_2\}, True,1.2 \right\rangle \}$
\end{center}
Since the formula of the WF is equivalent to True, we can evaluate this WF as $\exp(1.2 * |\Delta_{m_2}|)$.
\end{example}

\subsection{Caching}
The above rules each generate new MLNs, find their partition functions, combine and return the results. As we apply the above rules, we keep the partition functions of the generated MLNs in a cache so we can potentially use them in future when the partition function of the same MLN is required.

\section{Lifted Inference by Compiling into a Low-Level Program}
We explain \citet{LRC2CPP}'s LRC2CPP algorithm for compiling an MLN into a C++ program using an example. LRC2CPP is a recursive algorithm which takes as input an MLN $M$ and a variable name $vname$, and outputs a C++ code which computes $Z(M)$ and stores it in a variable called $vname$.

\begin{figure}[t]
\begin{center}
\includegraphics[width=\columnwidth]{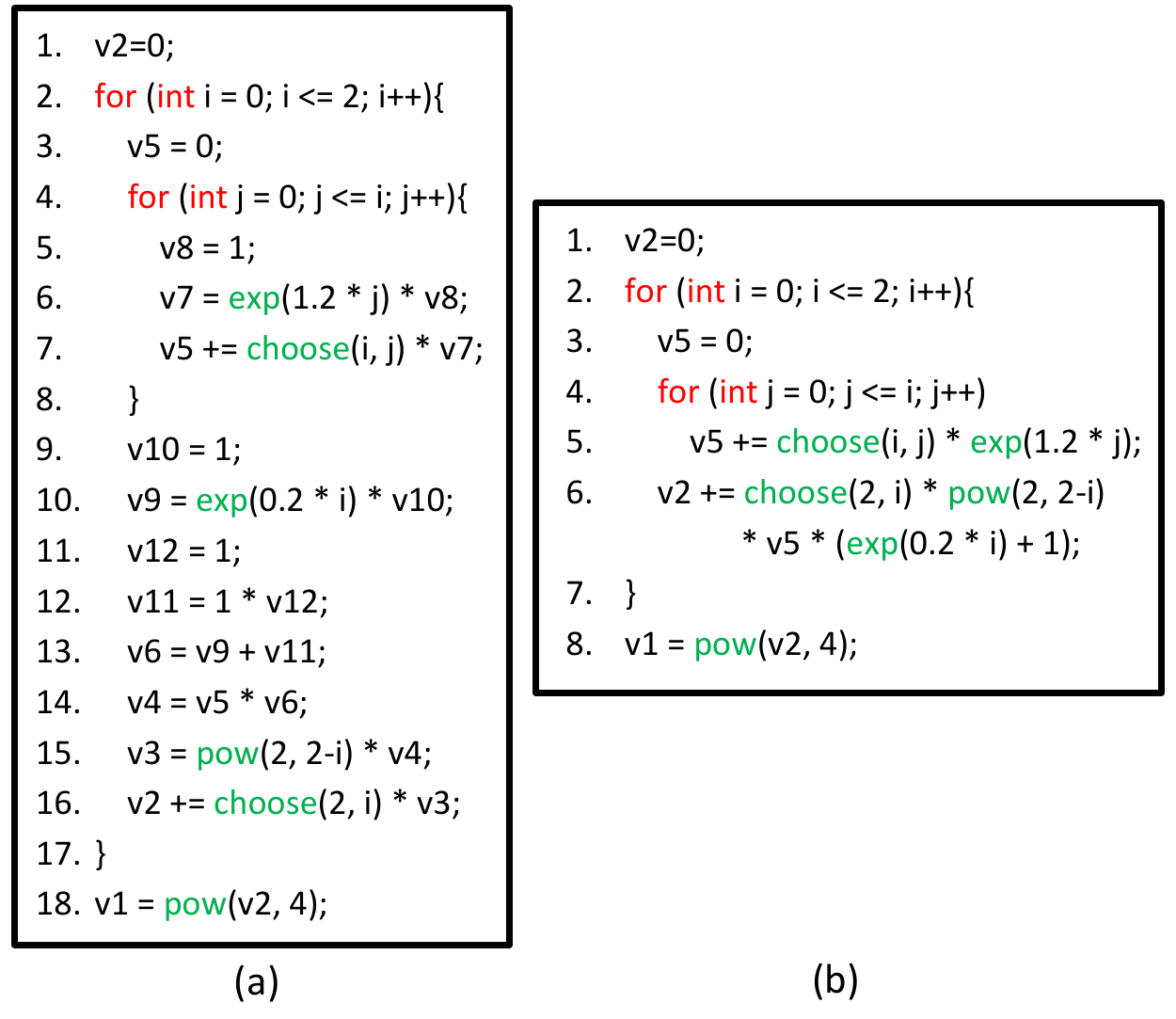}
\end{center}
\caption{(a) The C++ program for the MLN in Example~\ref{mln-example}. The partition function ($Z$) is stored in $v1$. (b) The C++ program in part (a) after pruning.}
\label{cpp}
\end{figure}

\begin{example}
Consider compiling the MLN $\mathbb{MLN}_1$ in Example~\ref{mln-example} to a C++ program by following LRC2CPP. 
Initially, LRC2CPP calls $LRC2CPP(\mathbb{MLN}_1,$\texttt{"}$v1$\texttt{"}$)$. 

As explained in Example~\ref{lifted-decomposition}, $x$ is a decomposer of $\mathbb{MLN}_1$. Let $\mathbb{MLN}_2$ denote $decompose(\mathbb{MLN}_1,x)$ (i.e. the resulting MLN in Example~\ref{lifted-decomposition} after decomposition). LRC2CPP generates the following C++ program: \\
\indent \indent $Code$ $for$ $LRC2CPP(\mathbb{MLN}_2,$\texttt{"}$v2$\texttt{"}$)$ \\
\indent \indent $v1=pow(v2, 5);$\\
where $5$ represents $|\Delta_x|$. For $LRC2CPP(\mathbb{MLN}_2,$\texttt{"}$v2$\texttt{"}$)$, suppose we choose to do a case analysis on $S(X_1,m)$ as in Example~\ref{lifted-case-analysis}. Assuming $\mathbb{MLN}_3$ represents $\mathbb{MLN}_2$ conditioned on $Obs(S(X_1, m), i)$, LRC2CPP generates a \emph{for loop} as follows:\\
\indent \indent $v2 = 0;$ \\
\indent \indent $for(int$ $i=0;$ $i <= 2;$ $i$++$)\{$\\
\indent \indent \indent $Code$ $for$ $LRC2CPP(\mathbb{MLN}_3,$\texttt{"}$v3$\texttt{"}$)$ \\
\indent \indent \indent $v2$ += $Choose(2, i) * v3;$ \\
\indent \indent $\}$ \\
where $2$ represents $|\Delta_m|$, and $Choose(2,i)$ computes $\binom{2}{i}$. $\mathbb{MLN}_3$ has the WFs in the resulting MLN of Example~\ref{lifted-case-analysis}.

The formulae of the second and fourth WFs are $False$ and will be removed from $\mathbb{MLN}_3$. However, as explained in Example~\ref{remove-false-formulae}, after removing these two WFs, $R(X_1,m_2)$ will be totally eliminated. Assuming $\mathbb{MLN}_4$ represents $\mathbb{MLN}_3$ after removing its second and fourth WFs, LRC2CPP generates:\\
\indent \indent $Code$ $for$ $LRC2CPP(\mathbb{MLN}_4, $\texttt{"}$v4$\texttt{"}$)$\\
\indent \indent $v3 = pow(2, 2-i) * v4;$\\
where $2-i$ refers to the number of ground variables in $R(X_1, m_2)$ (i.e. $|\Delta_{m_2}|$). 

$\mathbb{MLN}_4$ is disconnected as explained in Example~\ref{disconnected-components}. Let $\mathbb{MLN}_{41}$ and $\mathbb{MLN}_{42}$ represent the first and second connected components. LRC2CPP generates:\\
\indent \indent $Code$ $for$ $LRC2CPP(\mathbb{MLN}_{41},$\texttt{"}$v5$\texttt{"}$)$\\
\indent \indent $Code$ $for$ $LRC2CPP(\mathbb{MLN}_{42},$\texttt{"}$v6$\texttt{"}$)$\\
\indent \indent $v4 = v5 * v6;$\\
The first component requires a case analysis of a PRV with one logical variable which generates another for loop, and the second component requires a case analysis of a PRV with no logical variables. We can continue following LRC2CPP for these components and get the C++ program in Figure~\ref{cpp}(a).
\end{example}

\subsection{Optimizing C++ Programs}
Since the program obtained from LRC2CPP is generated automatically (not by a developer), a post-pruning step might seem required to reduce the size of the program. For instance, one can remove lines 11 and 12 of the program in Figure~\ref{cpp}(a) and replace line 13 with \texttt{"}$v6 = v9 + 1;$\texttt{"}. The same can be done for lines 5 and 9. One may also notice that some variables are set to some values and are then being used only once. For example in the program of Figure~\ref{cpp}(a), $v7$ and $v9$ are two such variables. The program can be pruned by removing these lines and replacing them with their values whenever they are being used. One can obtain the program in Figure~\ref{cpp}(b) by pruning the program in Figure~\ref{cpp}(a). Pruning can potentially save time and memory at run-time, but the pruning itself may be time-consuming.

\citet{LRC2CPP} use available optimization packages for C++ programs which optimize the code at compile time. In particular, they use the $-O3$ flag at compile time to optimize their generated programs before running them.

\section{Experiments and Results}
\citet{LRC2CPP} compared their end-to-end running times to those of WFOMC \citep{Van:2011} and \emph{probabilistic theorem proving (PTP)} \citep{PTP} on six benchmarks. By varying the population sizes of the logical variables for these benchmarks, they showed that LRC2CPP beats these two approaches for most population sizes, especially when the population sizes are large. WFOMC was the closest rival of LRC2CPP. A question which remained unanswered in \citet{LRC2CPP}'s experiments was to whether LRC2CPP outperforms WFOMC because the compilation to a target circuit is faster in LRC2CPP, or because reasoning with the target circuit generated by LRC2CPP is more efficient than that of WFOMC.

In order to address the above question, we measured the time spent by LRC2CPP and WFOMC on each of the reasoning steps for three networks: 1- the network used in Figure~1(f) of \citep{LRC2CPP}, 2- a network with only one WF $A(x)\wedge B(x) \wedge C(x,m) \wedge D(m) \wedge E(m) \wedge F$, and 3- another network with only with WF $A(x)\wedge B(x) \wedge C(x) \wedge D(x,m) \wedge E(m) \wedge F(m)  \wedge G(m) \wedge H$. For LRC2CPP, we used the MinNestedLoops heuristic \citep{LRC2CPP} to select the (lifted) case analysis order of PRVs. MinNestedLoops starts with the order obtained from MinTableSize \citep{Kazemi:2014a} and tries to improve it in terms of the \emph{maximum number of nested loops} it produces in the C++ program using stochastic local search.
All experiments were conducted on a 2.8GH core with 4GB RAM under MacOSX. Unless stated otherwise, the C++ programs of LRC2CPP were compiled using \texttt{g++} compiler.

\begin{figure*}[t]
\begin{center}
\includegraphics[width=0.85\textwidth]{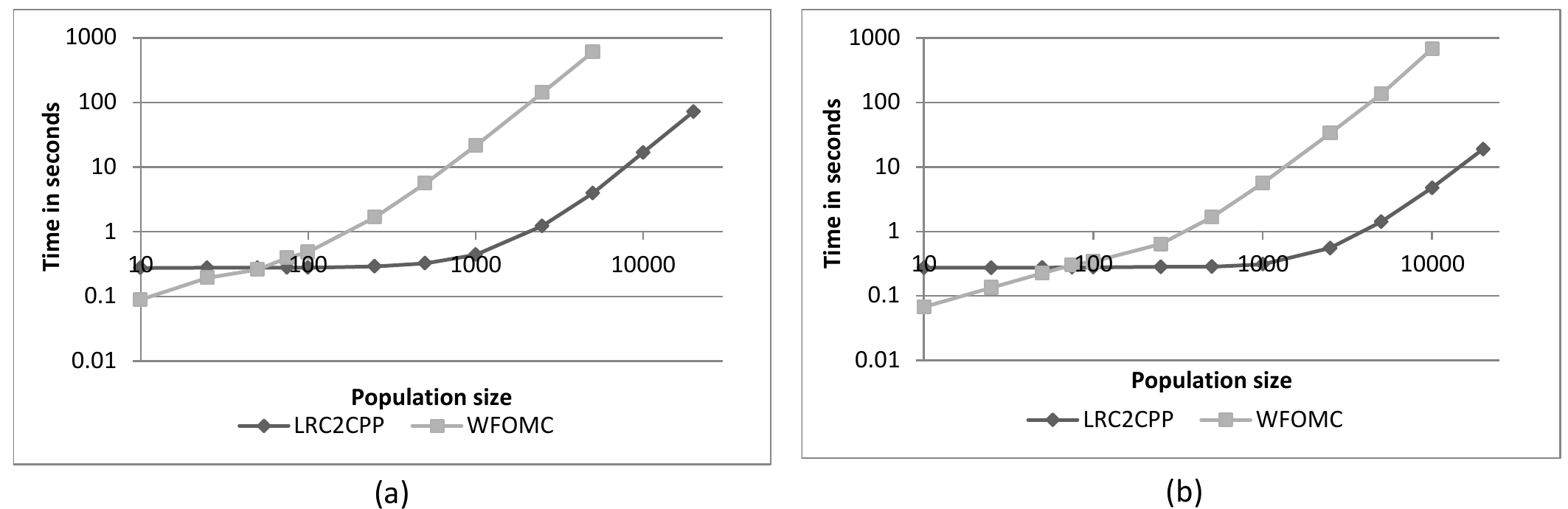}
\end{center}
\caption{The amount of time spent for reasoning with the target circuit in LRC2CPP and WFOMC on two benchmarks and for different population sizes.}
\label{compare-steps-fig}
\end{figure*}

For the three networks, it takes LRC2CPP 0.173, 0.029, and 0.138 seconds and it takes WFOMC 0.768, 0.373, and 0.512 seconds respectively to generate their target circuits. Figure~\ref{compare-steps-fig}(a),~(b) represent the time spent by LRC2CPP and WFOMC for reasoning with their circuits\footnote{Reasoning with LRC2CPP programs are considered as the time spent on compiling the C++ codes plus the run time.} for the first and second networks when the population of the logical variables varies at the same time (WFOMC could not solve the third circuit for population sizes $\geq 500$, so we did not include the run-time diagram for the third network).

\begin{figure*}
\begin{center}
\includegraphics[width=\textwidth]{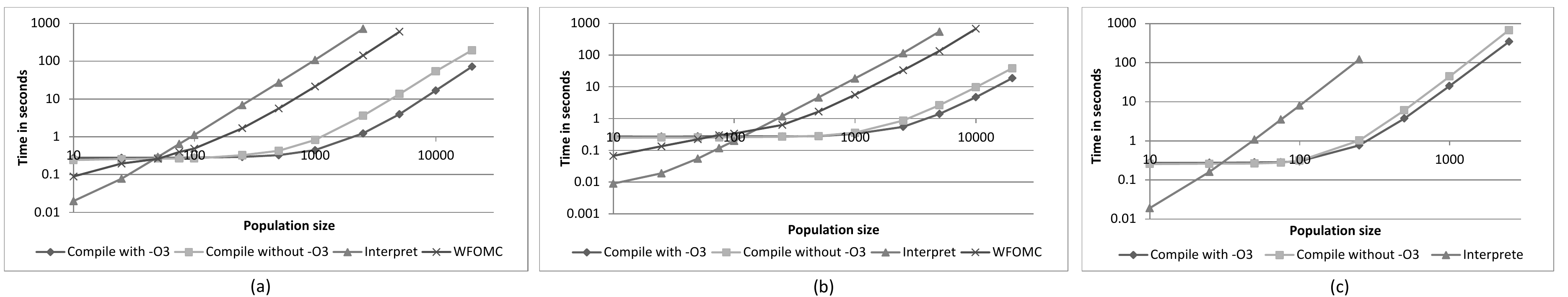}
\end{center}
\caption{The amount of time spent for reasoning with the programs generated by LRC2CPP when the program is interpreted, compiled, or compiled and optimized as well as the amount of time spent for reasoning with WFOMC's target circuit on three benchmarks and for different population sizes (WFOMC failed to produce an answer in less than $1000s$ for the third network when the population size was $500$ or higher, so we did not include it in the diagram).}
\label{compile-effect}
\end{figure*}

Obtained results represent that the compilation part takes almost the same amount of time in both LRC2CPP and WFOMC. For small population sizes, reasoning with WFOMC's circuit is more efficient because LRC2CPP's circuit needs a program compilation step. However, as the population size grows, the program compilation time becomes negligible and reasoning with the programs generated by LRC2CPP becomes much faster than reasoning with the data structures generated by WFOMC. As an example, it can be seen from the diagrams that reasoning with LRC2CPP's program offers about $163x$ speedup compared to WFOMC's data structure for the first network when the population sizes are $5000$. It is also interesting to note that the slope of the diagrams are the same, meaning reasoning with both circuits has the same time complexity.

Our first experiment indicates that the speedup in LRC2CPP is mostly due to the reasoning step. The next question to be answered is why reasoning with LRC2CPP's programs is more efficient than reasoning with WFOMC's data structures. \citet{LRC2CPP} hypothesized that the speedup is due to the fact that LRC2CPP's programs can be compiled and optimized, while reasoning with WFOMC's data structures requires an interpreter: a virtual machine that executes the data structure node-by-node. Validating this hypothesis by comparing the runtimes of LRC2CPP and WFOMC softwares is not sensible as there might be several implementation or other differences (e.g., case analysis order) between the two softwares.

In order to test \citet{LRC2CPP}'s hypothesis in an implementation-independent way, we used LRC2CPP to generate programs for the three networks in our previous experiment. For the reasoning step, we ran the programs in three different ways: 1- compiling and optimizing the programs using $-O3$ flag, 2- compiling without optimizing the programs, and 3- running the programs using Ch 7.5.3 which is an interpreter for C++ programs\footnote{Note that C++ interpreters are mostly used for teaching purposes and may not be highly optimized.}. Obtained results can be viewed in Figure~\ref{compile-effect}. We also included the run time of WFOMC in the first two diagrams (as explained before, WFOMC failed on the third network for population sizes of $500$ or more). It can be viewed that interpreting the C++ programs produces similar run times as working with WFOMC's data structures. The diagram for WFOMC is slightly below the diagram for interpreting the C++ program. One reason can be the non-optimality of the interpreter used for interpreting the C++ programs. It is interesting to note that by interpreting the C++ programs for small population sizes, and compiling and optimizing them for larger population sizes, in our benchmarks LRC2CPP's programs are always more efficient than the WFOMC's data structures.

In order to compare the speedup caused by compilation (instead of interpreting) with the speedup caused by optimization, we measured the percentage of speedup caused by each of them for our three benchmarks (we only considered the cases where both of them contributed to the speedup). We found that on average, 99.7\% of the speedup is caused by compilation, and only 0.3\% of it is caused by optimization. For the largest population where interpreting produced an answer in less than $1000s$, we found that compilation offers an average of \emph{175x} speedup compared to interpretation. Furthermore, for the largest population where compilation produced an answer, we found that optimization offers an average of \emph{2.3x} speedup compared to running the code without optimizing it.  

\section{Conclusion}
Compiling relational models into low-level programs for lifted probabilistic inference is a promising approach and offers huge speedups compared to the other approaches. In this paper, we conducted two experiments to explore the reasons behind the efficiency of this approach. In our first experiment, we compared compiling to low-level programs vs. WFOMC (which compiles into data structures) regarding the amount of time spent on different steps of the reasoning process. Our results indicated that the compilation step takes almost the same time in both approaches and almost all the speedup comes from reasoning with a low-level program instead of a data structure. In our second experiment, we explored why reasoning with a low-level program is more efficient than reasoning with a data structure. We designed an implementation-independent experiment using which we tested and validated \citet{LRC2CPP}'s hypothesis stating that low-level programs can be \emph{compiled} and \emph{optimized}, while reasoning with a data structure requires a virtual machine to \emph{interpret} the computations, and compilers are known to be faster than interpreters. 

\bibliography{MyBib}
\bibliographystyle{aaai}

\end{document}